\documentclass[letterpaper]{article} %
\usepackage[preprint]{aaai2027}
\usepackage[hyphens]{url}  %
\usepackage{graphicx} %
\usepackage{natbib}  %
\usepackage{caption} %
\usepackage{algorithm}
\usepackage{algorithmic}

\usepackage{newfloat}
\usepackage{listings}
\DeclareCaptionStyle{ruled}{labelfont=normalfont,labelsep=colon,strut=off} %
\floatstyle{ruled}
\newfloat{listing}{tb}{lst}{}
\floatname{listing}{Listing}

\usepackage{booktabs}

\title{CoSA: Accelerating Long-Context Inference via Proxy-Kernel\\Co-Designed Sparse Attention}
\author{
    Yufei Xue\textsuperscript{\rm 1, 2}\equalcontrib\thanks{Work done during an internship at Tencent.},
    Lin Niu\textsuperscript{\rm 1}\equalcontrib,
    Hong Liu\textsuperscript{\rm 1},
    Siran Liu\textsuperscript{\rm 1},
    Hanyong Shao\textsuperscript{\rm 1},
    Wei Liu\textsuperscript{\rm 1},\\
    Guanghua Yu\textsuperscript{\rm 1}\corresponding,
    Jianchen Zhu\textsuperscript{\rm 1},
    Jun Zhang\textsuperscript{\rm 2}\corresponding
}
\affiliations{
    \textsuperscript{\rm 1}Tencent
    \textsuperscript{\rm 2}The Hong Kong University of Science and Technology\\
    yufei.xue@connect.ust.hk; lucayu@tencent.com; eejzhang@ust.hk
}

\usepackage{amsmath, amsthm, amssymb}
\usepackage{bm}

\theoremstyle{plain}

\theoremstyle{definition}

\DeclareMathOperator*{\argmax}{arg\,max}
\DeclareMathOperator{\rowmax}{rowmax}
\DeclareMathOperator{\rowsum}{rowsum}

\usepackage{xspace}
\newcommand{\method}{CoSA\xspace}
\newcommand{\methodd}{CoSA$^{\circ}$\xspace}
\usepackage{booktabs, multirow, array}
\usepackage[table]{xcolor}
\definecolor{oursblue}{HTML}{0074BC} %
\definecolor{propcolor}{HTML}{2C5F9E}  %
\definecolor{limitcolor}{HTML}{A23B47}  %
\newcommand{\bg}{\rowcolor{gray!15}}
\newcommand{\bgii}{\rowcolor{gray!8}}
\newcommand{\bgiii}{\rowcolor{gray!3}}
\newcommand{\hlb}[1]{\textcolor{blue}{#1}}
\newcommand{\cmt}[1]{\textcolor{gray}{#1}}

\usepackage{pifont}
\newcommand{\cmark}{\ding{51}} %

\newtheoremstyle{propstyle}{}{}{\itshape}{0pt}{}{.}{.5em}{\bfseries\color{propcolor}\thmname{#1}\thmnumber{ #2}\thmnote{\ \normalfont\color{black}(#3)}}
\newtheoremstyle{limitstyle}{}{}{\itshape}{0pt}{}{.}{.5em}{\bfseries\color{limitcolor}\thmname{#1}\thmnumber{ #2}\thmnote{\ \normalfont\color{black}(#3)}}
\theoremstyle{propstyle}
\newtheorem{property}{Property}
\theoremstyle{limitstyle}
\newtheorem{limitation}{Limitation}
\newcommand{\prop}[1]{\textcolor{propcolor}{\textbf{P#1}}}
\renewcommand{\lim}[1]{\textcolor{limitcolor}{\textbf{L#1}}}

\begin{document}

\maketitle

\begin{abstract}
The quadratic cost of self-attention makes long-context inference prohibitively expensive, and proxy-based block-sparse attention has become a practical remedy. Existing methods typically rely on a proxy to predict a binary sparse mask and a kernel to consume this mask and perform sparse attention computation. Such an approach is effective under moderate budgets. However, as the budget tightens, the estimated proxy inevitably drops some salient blocks, while the kernel can only apply the sparse mask mechanically, leading to an evident drop in model accuracy. We propose CoSA, a two-stage training-free \underline{S}parse \underline{A}ttention under proxy-kernel \underline{CO}-design, which couples a Kernel-Aware Proxy (KAP) with an Ordered-Skipping Kernel (OSK). In the first stage, the KAP selects blocks under a \textit{moderate} budget and produces an \textit{ordered} mask that prescribes the order in which KV pages are visited in the kernel inner loop. In the second stage, the OSK applies this mask and skips more blocks under a tightened budget given online-softmax statistics. Across mainstream LLM backbones and long-context benchmarks, CoSA attains higher accuracy at lower budgets. Impressively, CoSA achieves a 4.93$\times$ attention speedup and reduces end-to-end Time-to-First-Token by 2.53$\times$ under a context length of 128K with negligible performance degradation. Code is available at https://github.com/Tencent/AngelSlim.
\end{abstract}

\section{Introduction}
\label{sec:intro}
Long context capabilities become the foundation for modern Large Language Models (LLMs), which are increasingly demanded by advanced use cases like Retrieval-Augmented Generation (RAG)~\cite{zheng2025retrieval,du2026rag} and autonomous agentic systems~\cite{team2025kimi-k2,zeng2026glm5}. However, the standard self-attention inherently incurs a computation cost that increases quadratically with respect to the sequence length, leading to unbearable inference latency for long inputs.

\begin{figure}[t]
    \centering
    \includegraphics[width=\linewidth]{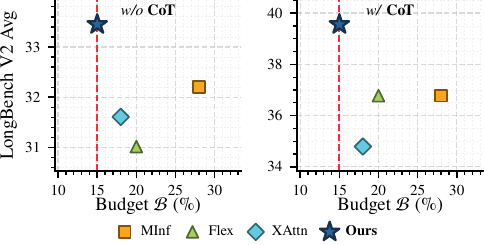}
    \caption{Performance and budget of mainstream sparse attention methods on LongBench-v2 with Qwen3-8B.}
    \label{fig:pareto_lb2}
\end{figure}

Sparse attention exploits the inherent sparsity of attention maps by selectively computing only the important query-key block interactions. Existing methods generally fall into training-free and trainable approaches. Trainable sparse attention learns which blocks to attend to via distillation~\cite{gao2024seerattention,zhao2026switch-attention,tang2026elastic-attention} or directly introduces sparse computation to pretraining. However, the training burden becomes problematic given the diverse array of model backbones, making training-based sparsification costly to deploy.

Training-free approaches instead decide block importance at inference time. The prevailing strategy uses a cheap proxy to estimate attention importance online, and it produces a binary sparse mask according to the sparsity budget.
The sparse attention kernel then follows this mask to perform the sparse attention computation~\cite{jiang2024minference,lai2025flexprefill,fan2026flashprefill,xu2025xattention,wang2025proxyattn}.
Such proxies are reliable under moderate budgets.
However, as the budget tightens, they increasingly miss the truly salient blocks.
There also exists a sparse strategy that skips blocks inside the kernel using the exact Softmax statistics~\cite{yuan2025blasst,zhang2025spargeattention}.
Though the skipping strategy is oracle-aligned, it still needs to compute the full query-key interactions. Moreover, it applies a conservative skip rule, which leaves much of the potential speedup on the table.

To reconcile this efficiency-fidelity trade-off, we propose \method, a two-stage training-free block-sparse attention for long-context inference built on the principle of \emph{proxy-kernel co-design}. Our design starts from inherent properties of block-sparse attention and limitations of in-kernel skipping (Sec.~\ref{sec:motivation}), which together argue for designing the proxy and kernel jointly rather than in isolation. The bridge between them is a single \emph{computation-order mask} that replaces the conventional binary block mask. Specifically, \method couples \textit{Kernel-Aware Proxy (KAP)} and \textit{Ordered-Skipping Kernel (OSK)}. The KAP prunes redundant dense query-key (QK) interactions under a moderate budget, which we call the first-stage sparsity.
Besides marking which blocks to compute, it also prescribes the order in which the selected blocks are visited in the block-sparse kernel loop.
The OSK then uses this computation-order mask via lightweight page-table remapping. On top of the computation-order mask, the OSK uses the exact in-kernel logits to skip even more blocks. This applies a second-stage sparsity that pushes the sparse budget lower. Overall, the proxy shapes the kernel by dictating its execution order, while the kernel shapes the proxy by consuming its mask to drive sparsity. This mutual shaping illustrates the principle of proxy-kernel co-design. Our contributions are as follows:

\begin{itemize}
    \item We revisit inherent properties of block-sparse attention and limitations of in-kernel skipping, revealing that the disconnection between the proxy and backend is the fundamental bottleneck to increasing sparsity while maintaining high accuracy.
    \item We design KAP, a proxy that emits a computation-order mask. It selects sparse blocks under a moderate budget, which realizes the first-stage sparse selection, and prescribes their visiting orders in the kernel. 
    \item We design OSK, an optimized kernel that echoes KAP by physically jumping over any-order pages based on the computation-order mask. It performs an in-kernel skip on softmax statistics, which carries out the second-stage sparse computation. 
    \item We conduct extensive experiments on mainstream LLM backbones and benchmarks, demonstrating superior accuracy at lower budgets. For example, \method delivers a 4.93$\times$ attention speedup and 2.53$\times$ end-to-end prefilling speedup at a 128K context length with negligible performance degradation.
\end{itemize}

\section{Related Works}
\label{sec:related}

\subsection{Sparse Attention}
Sparse attention skips unimportant query-key block interactions, and existing methods differ mainly in how the retained blocks are identified. MInference~\cite{jiang2024minference} estimates the optimal block indices at inference time from three pre-defined patterns, and FlexPrefill~\cite{lai2025flexprefill} extends it to a query-aware variant. XAttention~\cite{xu2025xattention} refines the pooling-based proxy with finer-grained antidiagonal scoring. Other works instead explore dynamic budget allocation~\cite{niu2026stem}, token-level sparsity~\cite{liu2026vecattention}, or head heterogeneity~\cite{wang2025proxyattn,liu2025unified,liu2026rrattention}. Effective as they are, all of these methods are inserted before a standard sparse-attention kernel and leave the kernel design itself untouched.

\subsection{High-Performance Attention Backend}
Beyond algorithm-level optimization, a parallel line of works optimizes the attention kernel itself. The FlashAttention series~\cite{dao2022flashattention,dao2023flashattention2,shah2024flashattention3,zadouri2026flashattention4} tile the computation and fuse the online-softmax (OSM) to cut traffic to GPU global memory, delivering large speedups without altering the result. Building on this, Block-Sparse-Attention kernel~\cite{guo2024block-sparse-attention} supports streaming and arbitrary block masks for efficient prefilling, and FlashInfer~\cite{ye2025flashinfer} provides a customizable engine with block-sparse and paged KV-cache formats. BLASST~\cite{yuan2025blasst} instead exploits the Softmax statistics to skip negligible blocks inside the kernel with near-zero decision overhead on modern GPUs. PagedAttention~\cite{kwon2023pageattention} stores the KV cache in non-contiguous pages to eliminate fragmentation during serving. Efficient as they are, these backends realize block sparsity by merely consuming a binary mask or depending on a conservative online-condition branch, which in turn constrains the algorithm design.

\begin{figure}[t]
    \centering
    \includegraphics[width=\linewidth]{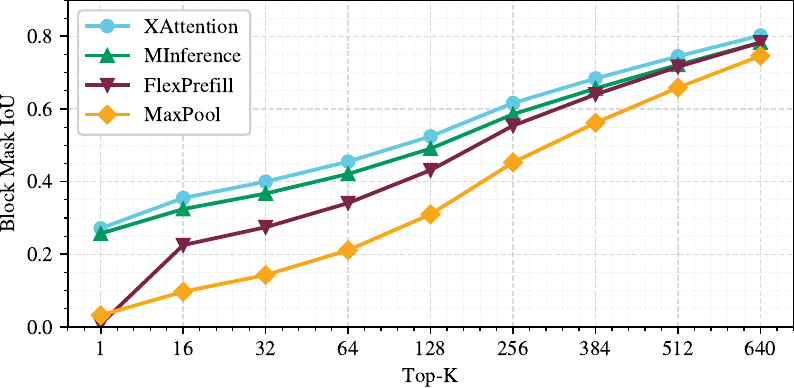}
    \caption{IoU of proxy masks against the oracle for Qwen3-8B under a context length of 128K.}
    \label{fig:proxy_oracle}
\end{figure}

\section{Motivation}
\label{sec:motivation}

\subsection{Dense Attention}
\label{sec:prelim}
FlashAttention~\cite{dao2023flashattention2} couples the tiling strategy with the OSM technique~\cite{milakov2018online-softmax} to compute standard attention. For the block-wise logits $\mathbf{S}_{ij}=\mathbf{Q}_i\mathbf{K}_j^\top$, the kernel maintains, per query row, a local and a running row-wise maximum (rowmax),
\begin{align}
\bm{m}^{\mathrm{loc}}_{ij}&=\rowmax(\mathbf{S}_{ij})\in\mathbb{R}^{b},\\
\bm{m}_{ij}&=\max\!\big(\bm{m}_{i,j-1},\,\bm{m}^{\mathrm{loc}}_{ij}\big)\in\mathbb{R}^{b},
\end{align}
where $b$ is the logical block size. The blocks in the kernel inner-loop are visited sequentially so that $\bm{m}_{ij}[r]=\max_{j'\le j}\bm{m}^{\mathrm{loc}}_{ij'}[r]$. Per block, the inner loop runs four steps: KV loading, logits computation, running rowmax update, and output rescaling. Building on this, we review two representative sparse-attention implementations.

\subsection{Mask-Driven Block Sparse Attention}
Mask-driven block-sparse attention uses a \textit{binary} block mask $\mathbf{M}$ to decide whether to compute or skip each QK block inside the block-sparse attention (BSA) kernel. The mask is derived from proxy attention scores $\mathbf{S}_\text{proxy} \in \mathbb{R}^{\lceil \frac{N}{b} \rceil \times \lceil \frac{N}{b} \rceil}$, defined as
\begin{equation}
\label{eq:proxy-mask}
\mathbf{M}[i,j]=\mathbf{1}\!\big[(i,j)\in\mathrm{TopK}_{\mathcal{B}}(\mathbf{S}_\text{proxy})\big] \in \{0,1\},
\end{equation}
where $N$ is the sequence length. A block with $\mathbf{M}[i,j]=0$ is dropped \emph{before} the kernel, so all four steps above are skipped for it. This saves the most per block but commits to an approximate mask. We point out the following two properties of mask-driven block sparse attention.

\begin{property}[Budget-dependent proxy fidelity]
Pre-kernel proxy masks (Eq.~\eqref{eq:proxy-mask}) track the truly important blocks well at a 
moderate budget, but lose the truly salient blocks under aggressive budgets.
\end{property}

\noindent A proxy is trustworthy only insofar as its mask recovers the blocks that genuinely matter. To quantify this, we treat the \emph{oracle mask} from \emph{full} attention as the ground truth and measure fidelity by the Intersection-over-Union (IoU) between each proxy mask (Eq.~\eqref{eq:proxy-mask}) and the oracle at a matched budget. Figure~\ref{fig:proxy_oracle} reports this IoU against the budget for mainstream proxies under the needle-in-a-haystack (NIAH)-style contexts. All variants track the oracle at moderate budgets but degrade as the budget tightens toward aggressive sparsity.

\begin{property}[Order-invariance of OSM]
FlashAttention accumulates each query block's output as a tiled OSM over its $J$ key/value blocks, and this result is independent of the order in which the blocks are visited. For any permutation $\rho_i$,
\begin{equation}
\label{eq:order-invariant}
\mathbf{O}_i=\mathcal{A}_i(1,2,\dots,J)=\mathcal{A}_i\big(\rho_i(1),\dots,\rho_i(J)\big),
\end{equation}
where $\mathcal{A}_i(\cdot)$ denotes the OSM computation of $i$-th query-block over the candidate key-blocks.
\end{property}

\noindent This order invariance allows candidate blocks to be visited in any order. Moreover, because modern serving frameworks store the KV cache in non-contiguous pages~\cite{kwon2023pageattention}, such arbitrary-order traversal can be implemented through lightweight KV-page remapping.

\subsection{In-Kernel Skip-Softmax}
Unlike the mask-driven BSA in the previous subsection, BLASST~\cite{yuan2025blasst} pioneers skipping Softmax logits within FlashAttention-4~\cite{zadouri2026flashattention4} to perform sparse computation. 
While computing each QK block's $\mathbf{S}_{ij}$ one by one, BLASST decides on the fly whether to skip a block, rather than relying on a proxy-based sparse mask, under the condition
\begin{align}
\label{eq:skip}
\mathrm{skip}(i,j)
&\Longleftrightarrow
\bigwedge_{r=1}^{b_q}\Big[\bm{m}^{\mathrm{loc}}_{ij}[r]-\bm{m}_{ij}[r]<\ln\frac{\Delta}{N}\Big]\notag\\
&\Longleftrightarrow
\max_{r}\big(\bm{m}^{\mathrm{loc}}_{ij}[r]-\bm{m}_{ij}[r]\big)<\ln\frac{\Delta}{N},
\end{align}
where the skip scale $\Delta$ adapts the threshold to the context length $N$. A row is negligible once its local max lies far below the running max. Under this condition, a block is skipped only when \emph{all} rows agree under the threshold $\Delta/N$. Unlike proxy estimation, which may lose information, this dense score $\mathbf{S}_{ij}$ is oracle-aligned and therefore accurate, at the price of a full QK multiplication per block.

Despite the fidelity of the in-kernel skip of Eq.~\eqref{eq:skip}, it suffers from the following two limitations.

\begin{limitation}[Dense $\mathbf{Q}_i\mathbf{K}_j^\top$ as proxy]
Because in-kernel skipping decides on the real logits, it must form $\mathbf{S}_{ij}=\mathbf{Q}_i\mathbf{K}_j^\top$ for every block. The QK multiplication is therefore unavoidable, and skipping trims only the value-side steps.
\end{limitation}

\noindent The savings of in-kernel skipping are thus inherently bounded. Because the decision is made on the real logits, the score must be materialized for every block before it can be judged. In-kernel skipping therefore trims the value-side work, while the quadratic query-key cost that dominates at long context is left untouched.

\begin{limitation}[Skip conservatism]
The skip condition of Eq.~\eqref{eq:skip} is conservative for two reasons:
\begin{enumerate}
\item (Running max $\neq$ global max) It tests each block against a running maximum that need not yet equal the global maximum;
\item (Bucket effect) Its per-row AND ($\bigwedge$) lets a single outlier row veto the skip of an otherwise negligible block.
\end{enumerate}
\end{limitation}

\begin{figure}[t]
\centering
\includegraphics[width=\linewidth]{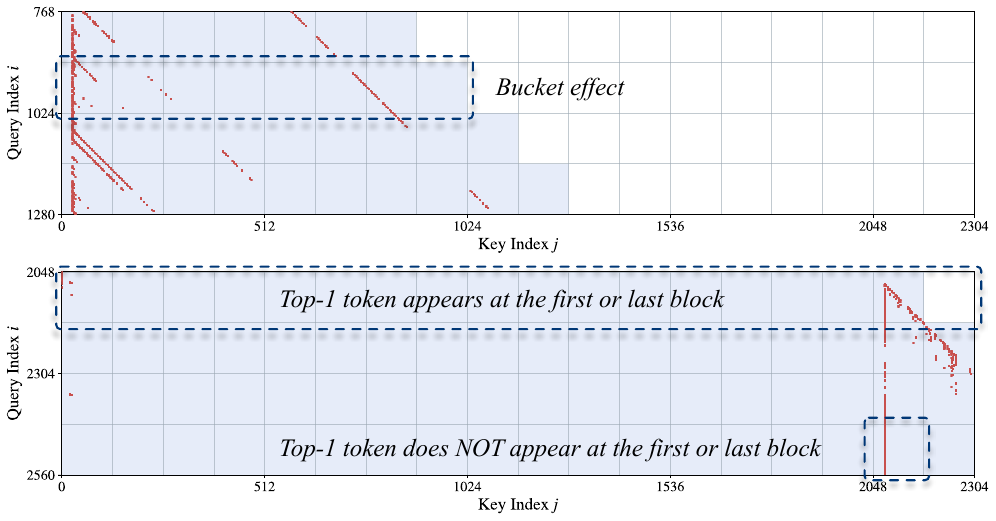}
\caption{Position of rowmax (red dots) over the attention map of two representative Qwen3-8B heads.}
\label{fig:top_stats}
\end{figure}

\paragraph{\lim{2.1} Running max $\neq$ global max.} The rule of Eq.~\eqref{eq:skip} tests against the running max $\bm{m}_{ij}$, which updated by only the \emph{already-visited} blocks $j'\le j$ and is a lower bound of the true global rowmax $\bm{m}^{\star}_{i}[r]=\max_{j'}\bm{m}^{\mathrm{loc}}_{ij'}[r]$. Since $\bm{m}_{ij}\le\bm{m}^{\star}_{i}$,
\begin{equation}
\bm{m}^{\mathrm{loc}}_{ij}[r]-\bm{m}_{ij}[r]\;\ge\;\bm{m}^{\mathrm{loc}}_{ij}[r]-\bm{m}^{\star}_{i}[r],
\end{equation}
so the criterion is strictly harder to satisfy than under the true global rowmax. As Figure~\ref{fig:top_stats} (bottom) shows, a rowmax can lie anywhere, not necessarily in the first (sink) or last (most-recent) block. Yet most high-performance backends still traverse KV blocks in a fixed ascending~\cite{nvidia2023tensorrtllm,zhang2024sageattention2,zhang2025sageattention2++} or descending~\cite{zadouri2026flashattention4,ye2025flashinfer,guo2024block-sparse-attention} order. This pre-defined order hinders redundant-block identification and thus yields suboptimal sparsity.

\begin{figure*}[t]
\centering
\includegraphics[width=\linewidth]{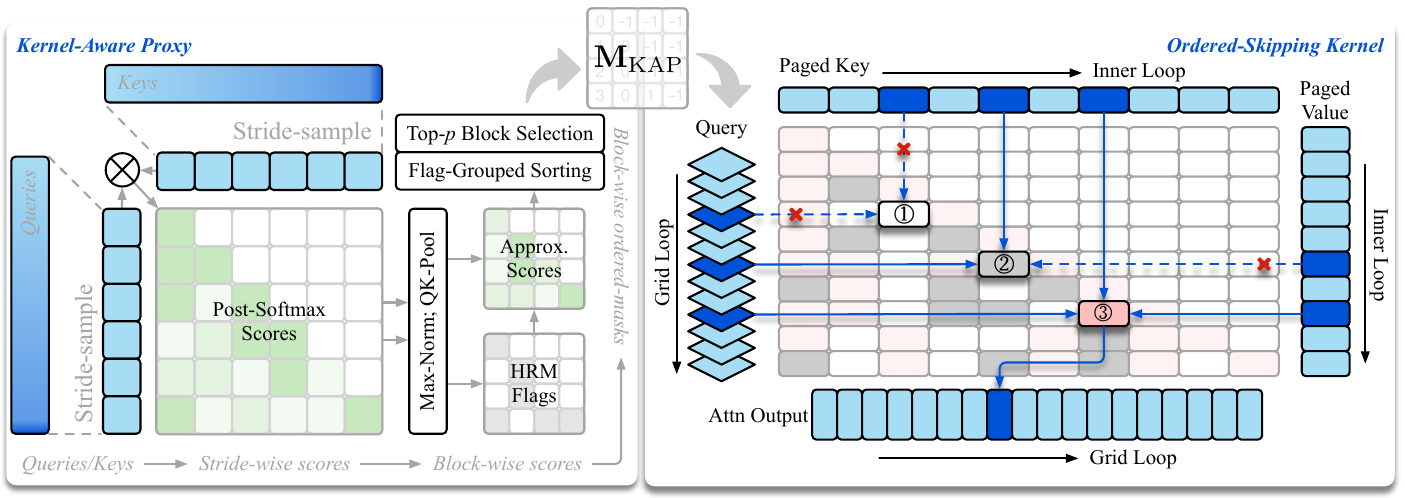}
\caption{Overview of \method. $\mathbf{M}_\text{KAP}$: the KAP mask serves as a bridge between the co-designed proxy and kernel. \textbf{Left:} KAP estimates a per-block score together with the HRM flag. They are merged into the computation-order mask $\mathbf{M}_\text{KAP}$, which places HRM blocks first. \textbf{Right:} OSK consumes $\mathbf{M}_\text{KAP}$ through KV page remapping, yielding three per-block patterns: \ding{192} mask-skipped, \ding{193} in-kernel skipped, and \ding{194} computed.}
\label{fig:overview}
\end{figure*}

\paragraph{\lim{2.2} Bucket effect.} Recall from the skip gate of Eq.~\eqref{eq:skip} that the AND ($\bigwedge$) over the $b_q$ rows lets the \emph{single worst} row
\begin{equation}
r^\star=\argmax_{r}\big(\bm{m}^{\mathrm{loc}}_{ij}[r]-\bm{m}_{ij}[r]\big)
\end{equation}
dictate the decision. Ideally, a single block $(m,n)$ contains the global rowmax for every query row. Visiting this block first would update the running maximum $\bm{m}_{mn}$ to the global maximum $\bm{m}^{\star}_{m}$. It allows every subsequent block to be evaluated against the exact global-max reference and skipped whenever it is truly negligible. However, these rowmax are scattered across blocks in practice, as shown in Figure~\ref{fig:top_stats} (top). Achieving the same oracle-aligned state therefore requires visiting all blocks that contain at least one row-wise maximum first. Until then, updating only $b-1$ rows is insufficient: the remaining row retains an underestimated running max and can veto the skip of an otherwise negligible block through the AND gate. This single-row restriction gives rise to the \textit{bucket effect}.

Overall, these properties and limitations directly motivate our design. At a moderate budget, the proxy is trustworthy (\prop{1}) and already filters part of the costly QK multiplication (\lim{1}). To reach an aggressive budget, we pair it with an exact in-kernel skip. Meanwhile, order-invariance (\prop{2}) frees us to reorder blocks and relieve the conservative skip (\lim{2}). We instantiate this as \method, described next.

\section{\method}
\label{sec:method}

\subsection{Overview}
Guided by the properties and limitations of Sec.~\ref{sec:motivation}, we propose \method, a two-stage sparse attention built on proxy-kernel co-design. \textit{On the proxy side}, to counteract \lim{2}, KAP leverages order-invariance (\prop{2}) to outputs an ordered mask $\mathbf{M}_\text{KAP}$ placing ``have rowmax'' (HRM) blocks first. Blocks filtered by $\mathbf{M}_\text{KAP}$ with moderate stage-1 budget $\mathcal{B}_\text{Stage-1}$ are directly skipped. \textit{On the kernel side}, OSK applies this mask and further skips value-side computation following Eq.~\eqref{eq:skip}, denoted as stage-2 sparse computation. As shown in Figure~\ref{fig:overview}, this yields three block types in OSK: \ding{192} mask-skipping, \ding{193} in-kernel skipping, and \ding{194} fully-computed. Algorithm~\ref{alg:jump} in Appendix~\ref{sec:method_appendix} summarizes the workflow.

\subsection{Kernel-Aware Proxy Design}
\label{sec:kap}
The KAP cheaply decides which blocks to retain and in what order ahead of the kernel. We design it directly around aggressive-sparsity failure (\prop{1}) and skip conservatism (\lim{2}). KAP emits a computation-order mask that prioritizes the visiting of HRM blocks under a moderate budget. Appendix~\ref{sec:method_appendix} provides an illustrative example in Figure~\ref{fig:illustrative_proxy}.

\paragraph{Block ranking.}
We first stride-downsample the queries and keys to $\mathrm{DS}(\mathbf{Q}) \in \mathbb{R}^{\frac{N}{s}\times D}$ and $\mathrm{DS}(\mathbf{K}) \in \mathbb{R}^{\frac{N}{s}\times D}$, where $s$ and $D$ are the stride size and hidden dimension, respectively. Then we form their approximate scores,
\begin{equation}
\hat{\mathbf{S}}=\text{Softmax}\left(\frac{\widetilde{\mathbf{Q}}\widetilde{\mathbf{K}}^\top}{\sqrt{d}}\right).
\end{equation}
Then we MaxPool over the keys of every block, and then max-normalize each sampled query row
\begin{align}
\text{MaxPool: }\mathbf{P}[r,j]&=\max_{c\in\mathcal{K}_j}\hat{\mathbf{S}}[r,c], \\
\text{MaxNorm: }\bar{\mathbf{P}}[r,j]&=\frac{\mathbf{P}[r,j]}{\max_{j'}\mathbf{P}[r,j']},
\end{align}
where $\mathcal{K}_j$ indexes the sampled keys of block $j$ and  $\bar{\mathbf{P}}\in \mathbb{R}^{\frac{N}{s}\times \frac{N}{b}}$ denotes the estimated scores aggregated in key-dimension. MaxPool preserves the rowmax within blocks, and MaxNorm turns exact estimated scores into relative ranking contribution, which are the direct counteractive design of \lim{2}. Over the \emph{queries}, we take the sum over sampled query rows $\mathcal{Q}_i$, letting every row of block $i$ contribute to the block score, which yields the estimated importance $\mathbf{S}^\text{KAP}_{ij}$ and the HRM flag $\mathbf{H}_{ij}$ simultaneously,
\begin{align}
\label{eq:score}
\mathbf{S}^\text{KAP}_{ij}&=\textstyle\sum_{r\in\mathcal{Q}_i}\bar{\mathbf{P}}[r,j], \\
\mathbf{H}_{ij}&=\mathbf{1}\!\big[\textstyle\max_{r\in\mathcal{Q}_i}\bar{\mathbf{P}}[r,j]=1\big]\in\{0,1\},
\end{align}
where $\mathbf{H} \in \mathbb{R}^{\frac{N}{b}\times \frac{N}{b}}$ is the HRM flags and $\mathbf{H}_{ij}$ marks that the query-block $i$ attains its rowmax inside key-block $j$.

\paragraph{Flag-grouped sorting.}
Fix a query-block $i$ with its $J=N/b$ candidate key-blocks, estimated scores $\mathbf{S}^\text{KAP}_i=(s_{ij})_{j=1}^{J}$ and HRM flags $\mathbf{H}_i=(h_{ij})_{j=1}^{J}$ from Eq.~\eqref{eq:score}. We turn these scores into a computation order $\rho_i$ by a flag-grouped sort that always prioritizes the HRM blocks:
\begin{equation}
\rho_i=\operatorname*{argsort}^{\downarrow}_{j\in\mathcal{H}_i} s_{ij}, \oplus \operatorname*{argsort}^{\downarrow}_{j\in\mathcal{R}_i} s_{ij},
\end{equation}
where $\operatorname*{argsort}^{\downarrow}_{\mathcal{X}}$ indicates sorting elements in $\mathcal{X}$ in a descending order. $\mathcal{H}_i=\{j: h_{ij}=1\}$ and $\mathcal{R}_i=\{j: h_{ij}=0\}$ denote the candidate blocks in the HRM group and the remainder. $\oplus$ concatenates two lists with the HRM group prioritized. We determine the moderate Stage-1 budget $\mathcal{B}_\text{Stage-1}$ via Top-$p$ selection following~\cite{xu2025xattention,lai2025flexprefill} and record each block's rank that yields the KAP mask
\begin{equation}
\label{eq:mkap}
\mathbf{M}_\text{KAP}[i,k] =
\begin{cases}
\rho_i[k], & 1\le k\le K_i \quad\text{($k$-th visited block),}\\
-1, & K_i<k\le J \quad\text{(padding),}
\end{cases}
\end{equation}
where $K_i=\mathcal{B}_\text{Stage-1}\times J$ is the number of selected blocks for the $i$-th query-block. In contrast to conventional proxies that yield a binary mask~\cite{jiang2024minference,lai2025flexprefill,xu2025xattention}, $\mathbf{M}_\text{KAP}$ encodes both \emph{which} blocks survive and \emph{in what order} they are visited.

\subsection{Ordered-Skipping Kernel}
\label{osk}
The OSK is the kernel-side half of our co-design, shaped by $\mathbf{M}_\text{KAP}$ so that the proxy's decisions become physical execution. It features three computation patterns: \ding{192} mask-skipped, \ding{193} in-kernel skipped, and \ding{194} computed, as shown in Figure~\ref{fig:overview}. We implement the OSK with the following designs.

\paragraph{Physical jumping over pages.}
Given $\mathbf{M}_\text{KAP}$ under a moderate budget (\prop{1}), the OSK partly skips the costly QK multiplication that in-kernel skipping alone can never save (\lim{1}). Recall that the Block-Sparse-Attention kernel~\cite{guo2024block-sparse-attention} traverses every block and drops the unselected ones through a logical mask, so its inner loop keeps paying the control-flow and synchronization cost even on skipped blocks. The OSK instead jumps directly to the next retained page and thereby removes these redundant synchronizations. Our jump is a page remap over the paged KV cache, and its visiting order is dictated by $\mathbf{M}_\text{KAP}$.
\paragraph{In-kernel logic skip.}
On top of the $\mathbf{M}_\text{KAP}$-driven skipping, the OSK further supports in-kernel skipping of Eq.~\eqref{eq:skip} to the blocks kept by stage-1 filtering. It realizes the stage-2 sparsification from the real online-softmax statistics rather than a proxy estimate. We implement this threshold-based skip following~\cite{yuan2025blasst,zhang2025spargeattention}, 
and further tailor its granularity to each 128-row query tile. Two consumer warp groups evaluate 64 query rows each and combine their partial predicates into a single tile-wide vote with negligible decision overhead.

\paragraph{Any-order page visiting.}
To counteract the conservative skipping rule (\lim{2}), the OSK visits pages in the order prescribed by KAP, which places the HRM blocks first so that the running maximum rises early and the subsequent in-kernel skip becomes both safer and more aggressive. Reordering the value accumulation leaves the output unchanged, as guaranteed by the order-invariance of the OSM (\prop{2}), so this reordering is correctness-preserving. It is realized by a cheap KV page remap: we feed the OSK directly with the \textit{activated block IDs list} (\textit{i.e.}, $\mathbf{M}_\text{KAP}$) emitted by KAP, which is already ordered. Because the remap operates at page granularity, the OSK can be seamlessly integrated into mainstream serving frameworks built on PagedAttention~\cite{kwon2023pageattention}.

\section{Experiments}
\label{sec:exp}

\begin{table}[t]
    \centering
    {\small\rmfamily
    \setlength{\tabcolsep}{1.6pt}
    \begin{tabular}{
    >{\raggedright\arraybackslash}m{1.2cm}
    *{6}{>{\centering\arraybackslash}c}
    >{\centering\arraybackslash}c
    >{\centering\arraybackslash}c
}
\toprule
\multirow{2}{*}{\textbf{Method}} & \multicolumn{6}{c}{\textbf{Context Length} (Acc, $\uparrow$)} & \multirow{2}{*}{\textbf{Avg.} ($\uparrow$)} & \multirow{2}{*}{\textbf{$\mathcal{B}$} ($\downarrow$)} \\
\cmidrule(lr){2-7}
& \textbf{4K} & \textbf{8K} & \textbf{16K} & \textbf{32K} & \textbf{64K} & \textbf{128K} & & \\
\midrule
\multicolumn{9}{c}{\textit{Qwen3-8B}} \\
\midrule
Dense & 95.46 & 94.20 & 93.95 & 92.57 & 83.72 & 75.30 & 89.20 & 100\% \\
MInf & 93.86 & 89.33 & \underline{91.83} & 92.04 & \textbf{83.49} & 70.09 & 86.77 & 49\% \\
Flex & \underline{94.62} & \underline{93.63} & 91.54 & \textbf{92.41} & 82.46 & \underline{71.68} & \underline{87.72} & 28\% \\
XAttn & 93.65 & 91.67 & 91.69 & 90.99 & 77.45 & 70.12 & 85.93 & \underline{24\%} \\
\bg \method & \textbf{95.25} & \textbf{94.25} & \textbf{93.34} & \underline{92.34} & \underline{82.88} & \textbf{73.84} & \textbf{88.65} & \textbf{22\%} \\
\midrule
\multicolumn{9}{c}{\textit{Llama-3.1-8B-Instruct}} \\
\midrule
Dense & 96.10 & 93.85 & 93.29 & 90.62 & 86.13 & 73.84 & 88.97 & 100\% \\
MInf & 94.02 & 93.47 & 93.00 & 90.20 & \textbf{86.01} & \underline{71.60} & \underline{88.05} & 47\% \\
Flex & 94.90 & \underline{94.01} & \underline{93.07} & 90.43 & 83.89 & 71.58 & 87.98 & \underline{29\%} \\
XAttn & \textbf{96.23} & \textbf{94.12} & \textbf{93.35} & \underline{90.75} & 82.56 & 70.40 & 87.90 & 33\% \\
\bg \method & \underline{95.73} & 93.59 & 93.02 & \textbf{90.82} & \underline{85.49} & \textbf{71.82} & \textbf{88.41} & \textbf{22\%} \\
\bottomrule
\end{tabular}

    }
    \caption{Main results (\%) on RULER. The best and second best results are \textbf{bolded} and \underline{underlined}, respectively. $\mathcal{B}$ is the weighted budget on sequence lengths.}
    \label{tab:ruler_compact}
\end{table}

\subsection{Setup}
\paragraph{Baselines and models.} We compare \method against strong block-sparse attention baselines, MInference~\cite{jiang2024minference}, FlexPrefill~\cite{lai2025flexprefill}, and XAttention~\cite{xu2025xattention}, with FlashAttention-2~\cite{dao2023flashattention2} as the full-attention baseline. We apply all methods to two representative backbones, Llama-3.1-8B-Instruct~\cite{grattafiori2024llama} and Qwen3-8B~\cite{yang2025qwen3}. For Qwen, we adopt the recommended sampling parameters and enable the thinking mode as each task requires, extending the context to 128K tokens with YaRN~\cite{peng2023yarn} where necessary.

\paragraph{Benchmarks.} We evaluate \method on widely recognized synthetic and real-world long-context benchmarks. RULER~\cite{hsieh2024ruler} provides synthetic tasks with token lengths ranging from 4K to 128K, while LongBench-v2 targets real-world deep understanding and reasoning. On LongBench-v2, we report both the direct (\textit{w/o} Chain-of-Thought, CoT) and the thinking (\textit{w/} CoT) modes, probing the reasoning robustness of sparse prefilling.

\paragraph{Implementation details.} Because the quadratic attention cost is concentrated in prefilling, our evaluation scope, in line with prior sparse-prefill studies~\cite{jiang2024minference,lai2025flexprefill,xu2025xattention}, sparsifies only the prefilling stage and leaves decoding dense. We implement the KAP scoring and pooling operations as a fused kernel to minimize proxy overhead. The OSK operates directly on the paged KV cache and supports page visits in the order prescribed by KAP. All accuracy and speed measurements are taken on an NVIDIA H20 node.

\paragraph{Hyperparameters.} For the stage-1 Top-$p$ selection, we fix $p=0.95$ for Qwen and use the offline-calibrated $p$-table for Llama. The scale factor $\Delta$ in the length-adaptive threshold $\Delta/N$ trades budget against accuracy. For example, we set $\Delta_\text{Qwen}=2000$ for \method. Full settings details are provided in Appendix~\ref{sec:add_exp}.

\begin{table}[t]
    \centering
    {\small\rmfamily
    \setlength{\tabcolsep}{3.5pt}
    \begin{tabular}{
    >{\raggedright\arraybackslash}m{1.2cm}
    *{4}{>{\centering\arraybackslash}m{0.9cm}}
    >{\centering\arraybackslash}m{0.9cm}
    >{\centering\arraybackslash}m{0.9cm}
}
\toprule
\multirow{2}{*}{\textbf{Method}} & \multicolumn{2}{c}{\textit{w/o} \textbf{CoT}} & \multicolumn{2}{c}{\textit{w/} \textbf{CoT}} & \multirow{2}{*}{\shortstack{\textbf{Avg.} ($\uparrow$)}} & \multirow{2}{*}{\shortstack{\textbf{$\mathcal{B}$} ($\downarrow$)}} \\
\cmidrule(lr){2-3} \cmidrule(lr){4-5}
& \textbf{Easy} & \textbf{Hard} & \textbf{Easy} & \textbf{Hard} & & \\
\midrule
\multicolumn{7}{c}{\textit{Qwen3-8B}} \\
\midrule
Dense & 39.19 & 32.72 & 44.53 & 36.82 & 37.48 & 100\% \\
MInf & \underline{36.98} & 29.26 & 43.23 & \underline{32.80} & \underline{34.49} & 28\% \\
Flex & 35.42 & 28.30 & \underline{44.27} & 32.15 & 33.90 & 20\% \\
XAttn & 31.25 & \textbf{31.83} & 40.10 & 31.51 & 33.20 & \underline{18\%} \\
\bg \method & \textbf{37.24} & \underline{31.11} & \textbf{45.83} & \textbf{35.69} & \textbf{36.51} & \textbf{15\%} \\
\midrule
\multicolumn{7}{c}{\textit{Llama-3.1-8B-Instruct}} \\
\midrule
Dense & 31.64 & 29.58 & 32.29 & 26.93 & 29.67 & 100\% \\
MInf & 30.21 & \textbf{30.87} & 30.21 & \textbf{31.19} & \underline{30.72} & 41\% \\
Flex & \underline{33.33} & 27.97 & \underline{31.25} & 28.30 & 29.72 & \underline{22\%} \\
XAttn & 29.17 & 27.01 & \textbf{32.29} & 29.26 & 29.13 & 35\% \\
\bg \method & \textbf{33.98} & \underline{28.94} & 30.99 & \underline{31.11} & \textbf{30.96} & \textbf{17\%} \\
\bottomrule
\end{tabular}

    }
    \caption{Main results (\%) on LongBench-v2 under both non-thinking (\textit{w/o} CoT) and thinking (\textit{w/} CoT) modes.}
    \label{tab:longbenchv2}
\end{table}

\subsection{Accuracy Results}

\paragraph{Ruler.}
Table~\ref{tab:ruler_compact} reports RULER accuracy from 4K to 128K on Qwen3-8B and Llama-3.1-8B. \method achieves the strongest average among the compared sparse baselines while using the lowest budget. More revealing than the averages is how this gap evolves with context length. At short contexts, every method sits close to full attention, so sparsity comes almost for free. However, as the sequence length increases, the baselines discard increasingly more salient blocks, resulting in substantial accuracy degradation. By 128K on Qwen3-8B, the strongest baselines have already slipped several points behind, while \method remains the strongest sparse method. The complete per-length results and additional operating points are provided in Appendix~\ref{sec:add_exp}.

\paragraph{LongBench-v2.}
Table~\ref{tab:longbenchv2} evaluates \method on the real-world deep-understanding and reasoning tasks of LongBench-v2, a more challenging setting than the synthetic retrieval of RULER. \method attains the best average among all sparse methods on both backbones while operating at the lowest budget, showing that aggressive sparsification need not come at the cost of reasoning quality. The advantage of \method holds under both evaluation modes. Whether the model answers directly (\textit{w/o} CoT) or reasons step by step (\textit{w/} CoT), \method remains the strongest sparse method, so its sparse prefilling does not disrupt the long CoT that the reasoning model depends on. 

\begin{figure*}[t]
\centering
\includegraphics[width=\linewidth]{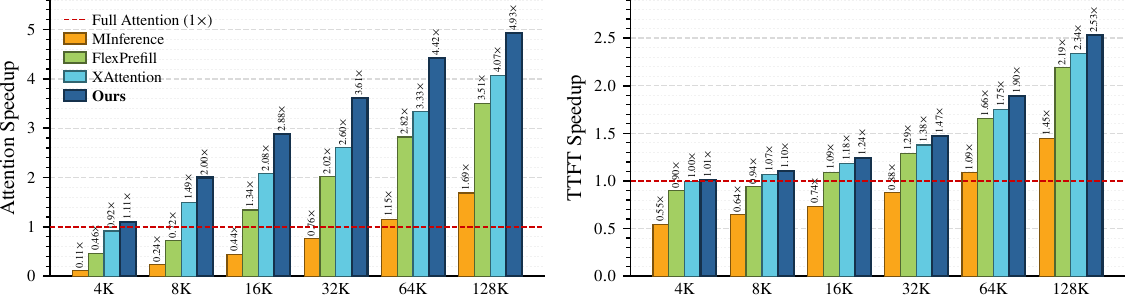}
\caption{Attention \textbf{(left)} and time-to-first-token (TTFT) \textbf{(right)} speedups over the Full Attention baseline across context lengths ranging from 4K to 128K.}
\label{fig:prefill_attention_speedup}
\end{figure*}

\begin{table}[t]
    \centering
    {\small\rmfamily
    \setlength{\tabcolsep}{0.2pt}
    \begin{tabular}{
    >{\raggedright\arraybackslash}m{2.0cm}
    *{2}{>{\centering\arraybackslash}m{1.0cm}}
    *{2}{>{\centering\arraybackslash}m{0.65cm}}
    >{\centering\arraybackslash}m{0.75cm}
    >{\centering\arraybackslash}m{1.1cm}
    >{\centering\arraybackslash}m{1.05cm}
}
\toprule
\multirow{2}{*}{\textbf{Variant}} & \multicolumn{2}{c}{\textbf{Proxy}} & \multicolumn{3}{c}{\textbf{Kernel}} & \multirow{2}{*}{\shortstack{\textbf{Avg.} ($\uparrow$)}} & \multirow{2}{*}{\shortstack{\textbf{$\mathcal{B}$} ($\downarrow$)}} \\
\cmidrule(lr){2-3} \cmidrule(lr){4-6}
& $\mathbf{M}_\text{01}$ & $\mathbf{M}_\text{KAP}$ & MS & IKS & RMP & & \\
\midrule
\multicolumn{8}{c}{\textit{Qwen3-8B}} \\
\midrule
Base & \cmark & & \cmark & & & 32.71 & 22\% \\
\bgiii \; \textit{+ KAP} & & \cmark & \cmark & & & \shortstack{33.27\\\textcolor{green!60!black}{(+0.56)}} & \shortstack{20\%\\\textcolor{green!60!black}{(-2\%)}} \\
\bgii \; \textit{+ IKS} & & \cmark & \cmark & \cmark & & \shortstack{32.27\\\textcolor{gray}{(-0.44)}} & \shortstack{16\%\\\textcolor{green!60!black}{(-6\%)}} \\
\bg \; \textit{+ RMP (Ours)} & & \cmark & \cmark & \cmark & \cmark & \shortstack{\textbf{33.45}\\\textcolor{green!60!black}{(+0.74)}} & \shortstack{\textbf{15\%}\\\textcolor{green!60!black}{(-7\%)}} \\
\midrule
\multicolumn{8}{c}{\textit{Llama-3.1-8B-Instruct}} \\
\midrule
Base & \cmark & & \cmark & & & 27.83 & 35\% \\
\bgiii \; \textit{+ KAP} & & \cmark & \cmark & & & \shortstack{29.67\\\textcolor{green!60!black}{(+1.84)}} & \shortstack{39\%\\\textcolor{gray}{(+4\%)}} \\
\bgii \; \textit{+ IKS} & & \cmark & \cmark & \cmark & & \shortstack{28.88\\\textcolor{green!60!black}{(+1.05)}} & \shortstack{24\%\\\textcolor{green!60!black}{(-11\%)}} \\
\bg \; \textit{+ RMP (Ours)} & & \cmark & \cmark & \cmark & \cmark & \shortstack{\textbf{30.86}\\\textcolor{green!60!black}{(+3.03)}} & \shortstack{\textbf{17\%}\\\textcolor{green!60!black}{(-18\%)}} \\
\bottomrule
\end{tabular}

    }
    \caption{Cumulative step-by-step ablation studies on LongBench-v2 (w/o CoT). The ``Base'' is naive mask-skipping (MS) via binary mask $\textbf{M}_{01}$. Variants are KAP, in-kernel skipping (IKS), and KV page remapping (RMP).}
    \label{tab:ablation}
\end{table}

\subsection{Efficiency Analysis}

\paragraph{Attention speedup.}
We first evaluate the attention speedup of \method against competing methods. We report speedups averaged over multiple runs across layers in Figure~\ref{fig:prefill_attention_speedup} (left). Two trends stand out. First, \method has already shown speedup at $1.11\times$ and is the only method to accelerate 4K-length context. Second, the advantage widens steadily as the sequence lengthens and the quadratic attention cost comes to dominate. Notably, \method climbs to $4.93\times$ at 128K and remains the leading method at every evaluated length. By comparison, the weakest baseline stays below full attention until 64K, underscoring how sensitive proxy-only methods are to selection overhead at short and moderate lengths.

\paragraph{End-to-end speedup.}
Figure~\ref{fig:prefill_attention_speedup} (right) shows that these attention gains carry through to the full prefilling stage. We evaluate this end-to-end efficiency using time-to-first-token (TTFT) latency. At 4K, \method introduces no regression. By 128K, it delivers a $2.53\times$ TTFT speedup while again leading all baselines. These results confirm that the computation saved by \method-enabled attention translates into a substantial reduction in real prefilling latency.

\subsection{Analysis}

\paragraph{Step-by-step ablation.}
To isolate the contribution of our co-design, Table~\ref{tab:ablation} reports a cumulative ablation on LongBench-v2 (\textit{w/o} CoT), starting from ``Base'' that uses an XAttention-style proxy. By replacing this proxy with our KAP, accuracy rises on both backbones at a comparable budget. It indicates that the ordered mask recovers salient blocks that a plain binary proxy tends to discard. By further turning on in-kernel skipping, it increases sparsity while incurring a small accuracy degradation on Qwen. The final step adds KV page remapping, leading to trend reversals. Accuracy is not merely restored but pushed clearly above the Base even under the tightest budget. Therefore, visiting the HRM blocks first makes the same in-kernel skip both safer and more aggressive. Overall, each component advances either accuracy or budget. Notably, remapping tightens budget without an accuracy degradation. This is exactly the interplay our co-design is built for, where the proxy-side ordering and the kernel-side skip only pay off when they act together.

\begin{figure}[t]
\centering
\includegraphics[width=\linewidth]{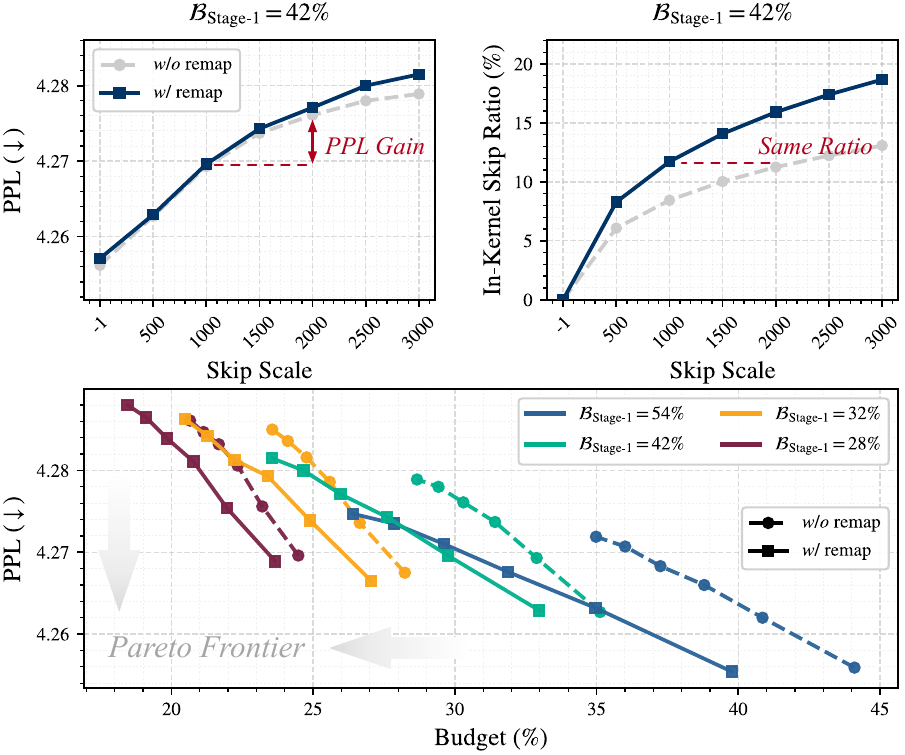}
\caption{Effect of KV page remapping on PPL and sparsity over 128 LongBench-v2 samples. \textbf{Top-left:} PPL varies smoothly with the skip scale $\Delta$. \textbf{Top-right:} remapping achieves a higher in-kernel skip ratio at comparable PPL. \textbf{Bottom:} across Stage-1 budgets, remapping forms the PPL-budget Pareto frontier.}
\label{fig:ppl_analysis}
\end{figure}

\paragraph{Effect of remapping.}
Figure~\ref{fig:ppl_analysis} analyzes the skip scale $\Delta$ and the benefit of KV page remapping. Using 128 random LongBench-v2 samples, we measure perplexity (PPL) and the in-kernel skip ratio across the Stage-1 budget $\mathcal{B}_\text{Stage-1}$, the skip scale $\Delta$, and the two remapping modes\footnote{The overall budget $\mathcal{B}$ is the Stage-1 budget $\mathcal{B}_\text{Stage-1}$ minus the in-kernel skip ratio. $\Delta=-1$ denotes no in-kernel skipping.}. The top-left panel first establishes that the skip scale is a well-behaved knob. As $\Delta$ shrinks, PPL decreases \textit{smoothly} and \textit{monotonically} toward the no-skip point. It implies $\Delta$ trades quality for sparsity in a predictable way and is safe to tune. The top-right panel then isolates what remapping does to the sparsity side. Evidently, we achieve a similar PPL level with more aggressive budgets. It echoes our co-designs: by front-loading the HRM blocks, more of the remaining blocks fall below the skip threshold without harming the output. Overall, \textit{w/} remap consistently attains lower PPL than \textit{w/o} at a matched skip ratio. Equivalently, a higher skip ratio at comparable PPL over the useful range of $\Delta$. Full results across $\mathcal{B}_\text{Stage-1}$ are given in Figure~\ref{fig:ppl_search}.

\paragraph{Frontier search.}
The bottom panel jointly plots PPL and the overall budget across several $\mathcal{B}_\text{Stage-1}$ settings. Each color fixes one Stage-1 budget, while the points along each curve sweep $\Delta$. For every Stage-1 budget from $54\%$ to $28\%$, remapping shifts the trajectory toward the lower-left region. Therefore, \method \textit{w/} remapping forms the PPL-budget Pareto frontier: at any target quality it reaches a lower budget, and at any target budget it reaches a lower PPL. The consistent shift across all four budgets confirms that the gain is not tied to a single Stage-1 operating point.

\section{Conclusion}
\label{sec:conclusion}
We presented \method, a two-stage training-free block-sparse attention for efficient long-context inference built on proxy-kernel co-design. Instead of treating the importance proxy and the sparse kernel as isolated designs, \method bridges them through a single computation-order mask. \method features KAP and OSK. The KAP selects blocks under a moderate budget and further prescribes the order in which they are visited. The OSK consumes this order through lightweight page remapping and skips additional blocks in-kernel from exact online-softmax statistics. The joint design lowers the budget while still retaining the truly salient blocks that an isolated proxy tends to discard. Across mainstream backbones and long-context benchmarks, \method attains higher accuracy at lower budgets than existing sparse methods. For example, it reduces TTFT by 2.53$\times$ at 128K context with negligible accuracy loss. As future work, we will extend \method beyond prefilling to decoding, which requires dedicated proxy and kernel designs due to its distinct query shapes and computational patterns.

\bibliography{aaai2027}

\clearpage
\appendix
\setcounter{figure}{0}
\setcounter{table}{0}
\setcounter{algorithm}{0}
\renewcommand{\thefigure}{\Alph{figure}}
\renewcommand{\thetable}{\Alph{table}}
\renewcommand{\thealgorithm}{\Alph{algorithm}}

\section{\method Appendix Overview}
\label{sec:outline_appendix}
This appendix is organized into three sections. Appendix Related Works extends the coverage of sparse-attention algorithms and high-performance backends. Additional Method Details provides a concrete KAP ordering example and the complete proxy-kernel workflow. Experimental Details and Additional Results document supplementary settings, operating points, and full accuracy-efficiency results.

\section{Related Works}
\label{sec:related_works_appendix}
\subsection{Training-Free Sparse Attention}
Training-free sparse prefilling relies on a lightweight proxy that predicts block importance before the attention kernel runs. MInference~\cite{jiang2024minference} and FlexPrefill~\cite{lai2025flexprefill} locate salient regions through vertical-slash (VS) and block-sparse pattern identification, while XAttention~\cite{xu2025xattention} captures both patterns by scoring each block along its antidiagonal. Several works instead refine the proxy itself: Stem~\cite{niu2026stem} augments it with value-aware selection and position-aware budget allocation, ProxyAttn~\cite{wang2025proxyattn} reduces its cost along the head dimension, and FlashPrefill~\cite{fan2026flashprefill} thresholds block scores to accommodate the long-tailed distribution of post-softmax logits. Beyond block-wise sparsity, VecAttention~\cite{liu2026vecattention} explores token-level sparsity to accelerate multimodal prefilling. A complementary line further extends sparsity to the decoding stage under long reasoning contexts~\cite{tang2024quest,wu2025tokenselect,luo2025asyncspade}.

\subsection{Distillation-based Sparse Attention}
Distillation-based sparsification keeps the backbone frozen and trains auxiliary modules that allocate sparsity dynamically. DuoAttention~\cite{xiao2024duoattention} and SwiAttn~\cite{zhao2026switch-attention} learn layer-wise routers that switch between dense attention and sliding-window attention (SWA), whereas Elastic Attention~\cite{tang2026elastic-attention} trains a task-aware router that grants a larger budget to harder tasks. RTTurbo~\cite{zhou2026rtturbo} exploits head-wise heterogeneity in sparsity to accelerate both prefilling and decoding, and SSA~\cite{shen2025ssa} aligns intermediate hidden features rather than performing end-to-end distillation for more efficient adaptation.

\subsection{Trainable Sparse Attention}
Beyond post-training adaptation, another line builds sparsity directly into the model during pretraining. NSA~\cite{yuan2025native-sparse-attention} and MoBA~\cite{lu2026moba} treat KV blocks as experts and activate only the salient ones at both the training and inference stages. DeepSeek-V3.2 refines NSA into finer-grained token-level selection, termed DeepSeek Sparse Attention (DSA), and DeepSeek-V4 further compresses attention through a hybrid of Compressed Sparse Attention (CSA) and Heavily Compressed Attention (HCA). HySparse~\cite{gao2026hysparse} interleaves dense and sparse layers and reuses the KV states of the dense layers, while MiniMax Sparse Attention (MSA) realizes trainable block-sparse attention on top of grouped-query attention (GQA).

\subsection{High-Performance Attention Backend}
Beyond algorithm-level optimization, a parallel line of work optimizes the attention kernel itself. The FlashAttention series~\cite{dao2022flashattention,dao2023flashattention2,shah2024flashattention3,zadouri2026flashattention4} tiles the computation and fuses the online softmax to cut traffic to GPU global memory, delivering large speedups without altering the result. SageAttention-2~\cite{zhang2024sageattention2} accelerates attention through nearly lossless INT4 quantization with outlier smoothing, while RingAttention~\cite{liu2023ringattention} distributes long sequences across multiple devices and overlaps the communication. MIT Block-Sparse-Attention~\cite{guo2024block-sparse-attention} supports streaming and arbitrary block masks for efficient prefilling. FlashInfer~\cite{ye2025flashinfer} offers a customizable engine with block-sparse and paged KV-cache formats, and PagedAttention~\cite{kwon2023pageattention} stores the KV cache in non-contiguous pages to remove fragmentation during serving. BLASST~\cite{yuan2025blasst} reuses the online-softmax statistics to skip negligible blocks inside the kernel with near-zero decision overhead on modern GPUs. 

\begin{table}[t]
    \centering
    {\small\rmfamily
    \setlength{\tabcolsep}{1.5pt}
    \begin{tabular}{
    >{\raggedright\arraybackslash}m{4.5cm}
    *{4}{>{\centering\arraybackslash}c}
}
\toprule
\multirow{2}{*}{\textbf{Backend}} & \multicolumn{2}{c}{\textbf{Mask-Skip}} & \multirow{2}{*}{\shortstack{\textbf{IKS}}} & \multirow{2}{*}{\shortstack{\textbf{P-RMP}}} \\
\cmidrule(lr){2-3}
& \shortstack{\textbf{Logic.}} & \shortstack{\textbf{Phy.}} & & \\
\midrule
MIT-BSA~\cite{guo2024block-sparse-attention} & \cmark & & & \\
FlashPrefill~\cite{fan2026flashprefill} & & \cmark & & \\
BLASST~\cite{yuan2025blasst} & & & \cmark & \\
SpargeAttn~\cite{zhang2025spargeattention} & & \cmark & \cmark & \\
\bg OSK (Ours) & & \cmark & \cmark & \cmark \\
\bottomrule
\end{tabular}

    }
    \caption{Comparison of sparse-attention backends.}
    \label{tab:kernel_comparison}
\end{table}

Table~\ref{tab:kernel_comparison} summarizes how representative backends realize sparse attention. MIT Block-Sparse-Attention performs logical mask skipping, whereas FlashPrefill and SpargeAttention physically jump to retained blocks. BLASST and SpargeAttention additionally support in-kernel skipping. Among the compared backends, OSK is the only one that combines physical page jumping, in-kernel skipping (IKS) based on exact logits, and any-order page visiting (P-RMP). This combination allows OSK to consume KAP's computation-order mask rather than a conventional binary mask.

\section{Additional Method Details}
\label{sec:method_appendix}

\subsection{Illustrative KAP Ordering}
\begin{figure}[t]
\centering
\includegraphics[width=0.9\linewidth]{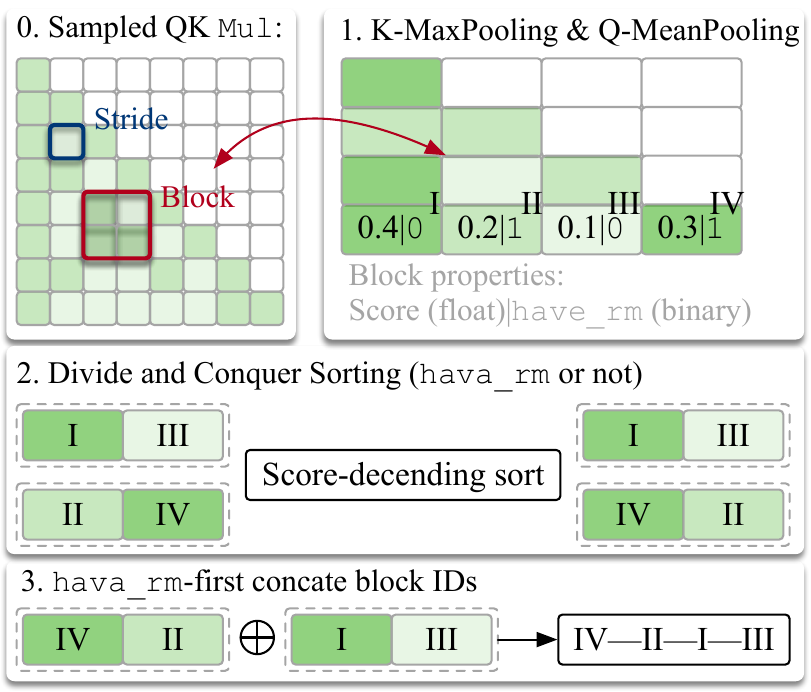}
\caption{HRM-first grouped ordering in KAP. Stride-downsampled scores are max-normalized and pooled into a per-block importance $s_{ij}$ and an HRM flag $h_{ij}$. Blocks are split by $h_{ij}$ and sorted by score within each group. Concatenating with the HRM group first gives the computation order in $\mathbf{M}_\text{KAP}$.}
\label{fig:illustrative_proxy}
\end{figure}

Figure~\ref{fig:illustrative_proxy} instantiates the flag-grouped ordering of Sec.~\ref{sec:kap} for one query block. KAP first assigns every candidate key block an importance score and an HRM flag. It then sorts the HRM and non-HRM groups independently by score, places the HRM group first, and retains the leading entries under the Stage-1 budget. The resulting sequence is written directly into $\mathbf{M}_\text{KAP}$, making both block selection and visiting order explicit.

\subsection{Complete Proxy-Kernel Workflow}
Algorithm~\ref{alg:jump} consolidates the two stages of \method. The first stage constructs $\mathbf{M}_\text{KAP}$ from the proxy scores and HRM flags. The second stage traverses only the retained KV pages in the prescribed order, updates the online-softmax statistics from exact logits, and skips the value-side computation whenever the in-kernel criterion is satisfied.

\begin{algorithm*}[t]
\caption{Complete Proxy-Kernel Workflow of \method}
\label{alg:jump}
\textbf{Input}: $\mathbf{Q}\in\mathbb{R}^{N\times d}$, Paged $\mathbf{K},\mathbf{V}\in\mathbb{R}^{N\times d}$\\
\textbf{Parameter}: Block sizes $b\times b$; stride $s$; \hlb{Stage-1 Top-$p$ threshold; skip scale $\Delta$}\\
\textbf{Output}: Output $\mathbf{O}$
\begin{algorithmic}[1]
\STATE \cmt{// Pre-kernel proxy (KAP), see Eqs.~\eqref{eq:score} and~\eqref{eq:mkap}}
\STATE \hlb{$\widetilde{\mathbf{Q}},\widetilde{\mathbf{K}}\gets\mathrm{Stride}(\mathbf{Q}),\mathrm{Stride}(\mathbf{K})$;\ \ $\hat{\mathbf{S}}\gets\text{Softmax}\!\big(\mathrm{CausalMask}(\widetilde{\mathbf{Q}}\widetilde{\mathbf{K}}^\top/\sqrt{d})\big)$} \COMMENT{Downsampling and cheap scores}
\STATE \hlb{$\mathbf{P}[r,j]\gets\max_{c\in\mathcal{K}_j}\hat{\mathbf{S}}[r,c]$;\ \ $\bar{\mathbf{P}}[r,j]\gets\mathbf{P}[r,j]/\max_{j'}\mathbf{P}[r,j']$} \COMMENT{Key-dim MaxPooling then max-norm}
\STATE \hlb{$s_{ij}\gets\textstyle\sum_{r\in\mathcal{Q}_i}\bar{\mathbf{P}}[r,j]$;\ \ $h_{ij}\gets\mathbf{1}[\max_{r\in\mathcal{Q}_i}\bar{\mathbf{P}}[r,j]{=}1]$} \COMMENT{Block score and HRM flag}
\STATE \hlb{$\rho_i\gets$ HRM-first grouped sort over causal candidates $j\le i$; apply Stage-1 Top-$p$ selection} \COMMENT{Ordered retained block IDs}
\STATE \cmt{// Attention loop}
\FOR{$i=1$ to $\lceil N/b\rceil$}
  \STATE $m_i\gets-\infty$,\ $l_i\gets 0$,\ $\mathbf{O}_i\gets\mathbf{0}$
  \STATE \cmt{// Traverse only the selected KV pages in $\mathbf{M}_\text{KAP}$}
  \FOR{\hlb{$k=1$ to $|\rho_i|$}}
    \STATE \hlb{$j\gets\rho_i[k]$; load pages $\mathbf{K}_j,\mathbf{V}_j$ via remap} \COMMENT{Block IDs remapping}
    \STATE $\mathbf{S}_{ij}\gets\mathbf{Q}_i\mathbf{K}_j^\top/\sqrt{d}$;\ \ $\bm{m}^{\mathrm{loc}}\gets\rowmax(\mathbf{S}_{ij})$
    \STATE $\bm{m}_i'\gets \bm{m}_i$;\ \ $\bm{m}_i\gets\max(\bm{m}_i,\bm{m}^{\mathrm{loc}})$
    \IF{\hlb{$\Delta>0\ \text{and}\ \max_r\big(m^{\mathrm{loc}}[r]-m_i[r]\big) <\ln\frac{\Delta}{N}$}}
      \STATE \hlb{\textbf{continue}} \COMMENT{In-kernel skip thresholding}
    \ENDIF
    \STATE $\widetilde{\mathbf{P}}_{ij}\gets\text{exp}(\mathbf{S}_{ij}-\bm{m}_i)$
    \STATE $l_i\gets \text{exp}({\bm{m}_i'-\bm{m}_i})l_i+\rowsum(\widetilde{\mathbf{P}}_{ij})$
    \STATE $\mathbf{O}_i\gets\mathrm{diag}(e^{\bm{m}_i'-\bm{m}_i})\mathbf{O}_i+\widetilde{\mathbf{P}}_{ij}\mathbf{V}_j$
  \ENDFOR
  \STATE $\mathbf{O}_i\gets\mathrm{diag}(l_i)^{-1}\mathbf{O}_i$; write $\mathbf{O}_i$
\ENDFOR
\STATE \textbf{return} $\mathbf{O}$
\end{algorithmic}
\end{algorithm*}

\section{Experimental Details and Additional Results}
\label{sec:add_exp}

\subsection{Experimental Configurations}\label{sec:exp_config_appendix}

\paragraph{\method settings.}
We use logical query, key, and value blocks of $b=128$ tokens. The physical KV-cache page size $b_p$ is configured independently by the paged-KV layout; a logical key/value block is resolved through one or more physical pages when $b_p$ differs from $b$.
Our SM90 kernel adopts a warp-specialized organization. Two consumer warp groups jointly process one 128-row query tile, with each consumer warp group responsible for 64 query rows. Each consumer warp group first reduces its local row-wise skip predicates, after which OSK combines the two partial results into a single tile-wide skip decision. This realizes the per-128 skip semantics of Eq.~\eqref{eq:skip}. The reduction directly reuses the on-chip row-wise maxima maintained by online softmax and introduces no additional score computation or global-memory traffic.

For Qwen3-8B, Stage-1 selection uses Top-$p$ with $p=0.95$, and the main operating point uses $\Delta=2000$ in the length-adaptive threshold $\Delta/N$. For Llama-3.1-8B, the Stage-1 $p$ values are calibrated offline by context length following~\cite{xu2025xattention}, and we set $\Delta=200$. Setting $\Delta=-1$ disables in-kernel skipping and serves as the Stage-1-only endpoint in the remapping analysis.

\subsection{Accuracy-Prioritized Operating Point}
The scale factor $\Delta$ controls the accuracy-budget trade-off of the in-kernel skip. In addition to the main \method setting reported in the body, we evaluate the accuracy-prioritized variant \methodd using a smaller $\Delta=1000$ for the Qwen model and $\Delta=200$ for the Llama model. As shown in Table~\ref{tab:ruler}, \methodd improves average RULER accuracy at a moderately higher budget, complementing the lower-budget operating point of \method. The table reports the complete per-length accuracy and realized budget for both backbones.

\begin{table*}[t]
    \centering
    {\small\rmfamily
    \setlength{\tabcolsep}{1.1pt}
    \begin{tabular}{
    >{\raggedright\arraybackslash}m{1.3cm}
    *{14}{c}
}
\toprule
\multirow{2}{*}{\textbf{Method}}
 & \multicolumn{2}{c}{\textbf{4K}} & \multicolumn{2}{c}{\textbf{8K}} & \multicolumn{2}{c}{\textbf{16K}}
 & \multicolumn{2}{c}{\textbf{32K}} & \multicolumn{2}{c}{\textbf{64K}} & \multicolumn{2}{c}{\textbf{128K}}
 & \multicolumn{2}{c}{\textbf{Avg.}} \\
\cmidrule(lr){2-3} \cmidrule(lr){4-5} \cmidrule(lr){6-7} \cmidrule(lr){8-9} \cmidrule(lr){10-11} \cmidrule(lr){12-13} \cmidrule(lr){14-15}
 & \textbf{Acc} ($\uparrow$) & \textbf{$\mathcal{B}$} ($\downarrow$) & \textbf{Acc} ($\uparrow$) & \textbf{$\mathcal{B}$} ($\downarrow$) & \textbf{Acc} ($\uparrow$) & \textbf{$\mathcal{B}$} ($\downarrow$) & \textbf{Acc} ($\uparrow$) & \textbf{$\mathcal{B}$} ($\downarrow$) & \textbf{Acc} ($\uparrow$) & \textbf{$\mathcal{B}$} ($\downarrow$) & \textbf{Acc} ($\uparrow$) & \textbf{$\mathcal{B}$} ($\downarrow$) & \textbf{Acc} ($\uparrow$) & \textbf{$\mathcal{B}$} ($\downarrow$) \\
\midrule
\multicolumn{15}{c}{\textit{Qwen3-8B}} \\
\midrule
Dense & 95.46 & 100\% & 94.20 & 100\% & 93.95 & 100\% & 92.57 & 100\% & 83.72 & 100\% & 75.30 & 100\% & 89.20 & 100\% \\
MInf & 93.86 & 71\% & 89.33 & 55\% & 91.83 & 49\% & 92.04 & 49\% & \textbf{83.49} & 60\% & 70.09 & 43\% & 86.77 & 49\% \\
Flex & 94.62 & \underline{49\%} & 93.63 & 45\% & 91.54 & 41\% & \underline{92.41} & 35\% & 82.46 & 29\% & 71.68 & 22\% & 87.72 & 28\% \\
XAttn & 93.65 & 53\% & 91.67 & 42\% & 91.69 & \underline{34\%} & 90.99 & \underline{26\%} & 77.45 & \underline{24\%} & 70.12 & \underline{21\%} & 85.93 & \underline{24\%} \\
\bg \method & \underline{95.25} & \textbf{36\%} & \underline{94.25} & \textbf{35\%} & \textbf{93.34} & \textbf{30\%} & 92.34 & \textbf{23\%} & 82.88 & \textbf{21\%} & \underline{73.84} & \textbf{20\%} & \underline{88.65} & \textbf{22\%} \\
\bg \methodd & \textbf{95.46} & 65\% & \textbf{94.41} & \underline{38\%} & \underline{93.08} & 35\% & \textbf{92.68} & \underline{26\%} & \underline{83.41} & \underline{24\%} & \textbf{73.96} & 26\% & \textbf{88.83} & 27\% \\
\midrule
\multicolumn{15}{c}{\textit{Llama-3.1-8B-Instruct}} \\
\midrule
Dense & 96.10 & 100\% & 93.85 & 100\% & 93.29 & 100\% & 90.62 & 100\% & 86.13 & 100\% & 73.84 & 100\% & 88.97 & 100\% \\
MInf & 94.02 & 75\% & 93.47 & 84\% & 93.00 & 84\% & 90.20 & 77\% & \textbf{86.01} & 55\% & 71.60 & 28\% & 88.05 & 47\% \\
Flex & 94.90 & \underline{48\%} & \underline{94.01} & 43\% & 93.07 & \underline{38\%} & 90.43 & \underline{32\%} & 83.89 & \underline{28\%} & 71.58 & 26\% & 87.98 & \underline{29\%} \\
XAttn & \textbf{96.23} & 70\% & \textbf{94.12} & 60\% & \textbf{93.35} & 50\% & 90.75 & 40\% & 82.56 & 32\% & 70.40 & 27\% & 87.90 & 33\% \\
\bg \method & 95.73 & \textbf{40\%} & 93.59 & \textbf{31\%} & 93.02 & \textbf{30\%} & \textbf{90.82} & \textbf{21\%} & 85.49 & \textbf{21\%} & \underline{71.82} & \textbf{21\%} & \underline{88.41} & \textbf{22\%} \\
\bg \methodd & \underline{95.93} & \underline{48\%} & 93.77 & \underline{41\%} & \underline{93.34} & 42\% & \underline{90.81} & 42\% & \underline{85.62} & 36\% & \textbf{72.92} & \underline{25\%} & \textbf{88.73} & 32\% \\
\bottomrule
\end{tabular}

    }
    \caption{Complete RULER results across context lengths. Each context length reports accuracy and realized computation budget for Qwen3-8B and Llama-3.1-8B-Instruct.}
    \label{tab:ruler}
\end{table*}

\subsection{Kernel Microbenchmark}
\begin{figure}[t]
\centering
\includegraphics[width=\linewidth]{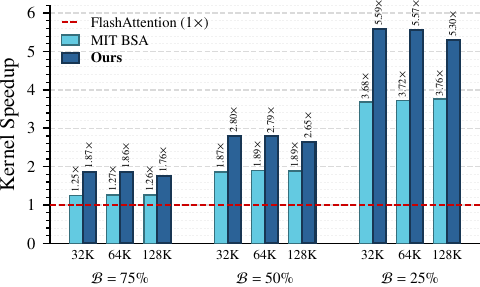}
\caption{Kernel-level speedup over FlashAttention~\cite{dao2023flashattention2} at budgets $\mathcal{B}\in\{25\%,50\%,75\%\}$ and context lengths $\{32\text{K},64\text{K},128\text{K}\}$. Our ordered-skipping kernel (index-driven physical jumping) is compared against the logical-skipping MIT BSA~\cite{guo2024block-sparse-attention}.}
\label{fig:kernel_speedup}
\end{figure}

Figure~\ref{fig:kernel_speedup} isolates the execution benefit of OSK from the proxy by comparing its physical page jumping with the logical skipping of Block-Sparse-Attention~\cite{guo2024block-sparse-attention} at matched budgets. The advantage grows as the context length increases and the retained budget decreases, because OSK removes the loop-control and synchronization work associated with masked iterations.

\subsection{Full Remapping Sweep Across Stage-1 Budgets}
\begin{figure*}[t]
\centering
\includegraphics[width=\linewidth]{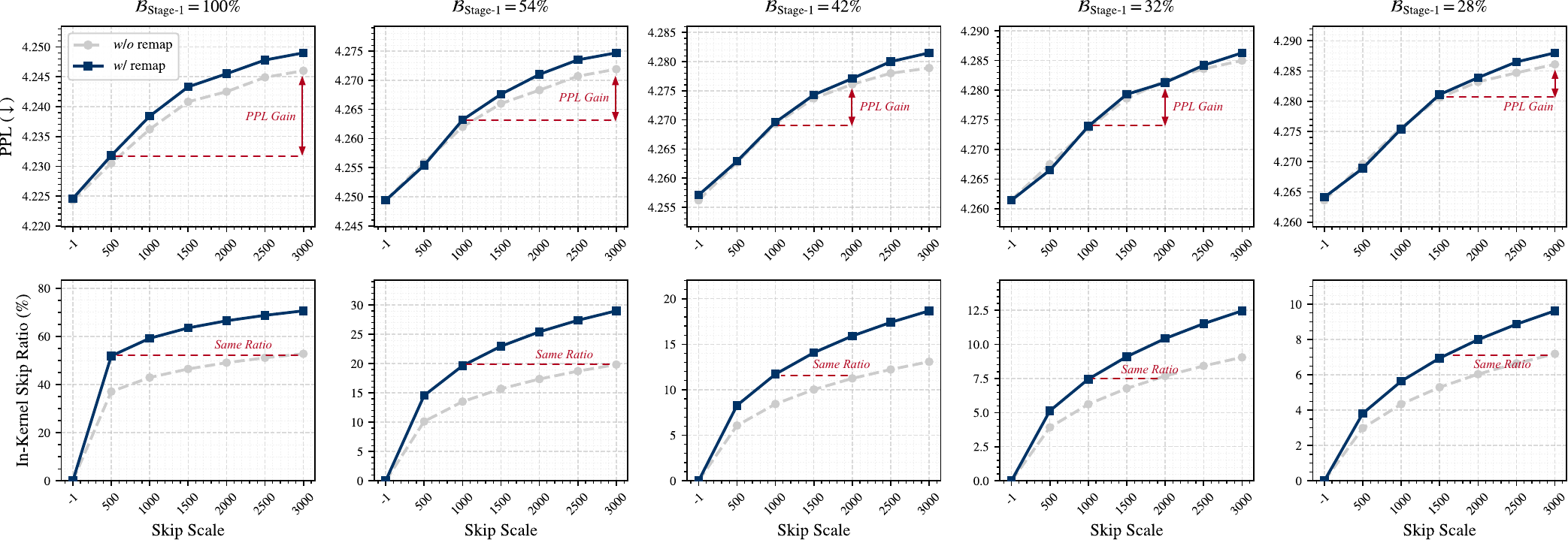}
\caption{Full remapping sweeps across Stage-1 budgets and skip scales. Each column fixes $\mathcal{B}_\text{Stage-1}$. The \textbf{Top} row reports PPL and the \textbf{Bottom} row reports the in-kernel skip ratio as $\Delta$ varies. Remapping achieves a higher skip ratio at a fixed $\Delta$ and a lower PPL at a matched skip ratio. Red annotations highlight the matched-ratio PPL gain.}
\label{fig:ppl_search}
\end{figure*}

Figure~\ref{fig:ppl_search} expands the remapping analysis across $\mathcal{B}_\text{Stage-1}\in\{100\%,54\%,42\%,32\%,28\%\}$. Each column fixes one Stage-1 budget: the top panel plots PPL against the skip scale $\Delta$, and the bottom panel reports the corresponding in-kernel skip ratio. Across all budgets, increasing $\Delta$ raises the skip ratio but also increases PPL, showing a smooth quality-sparsity trade-off. At a fixed $\Delta$, remapping consistently skips more blocks than the non-remapped execution. More importantly, remapping reaches a matched skip ratio with a smaller $\Delta$ and therefore a lower PPL, as indicated by the red annotations. For example, at $\mathcal{B}_\text{Stage-1}=100\%$, remapping reaches an approximately $52\%$ skip ratio at $\Delta=500$, whereas the non-remapped execution requires roughly $\Delta=3000$ for a similar ratio and incurs a higher PPL. Although the absolute in-kernel skip ratio decreases as the Stage-1 budget tightens, the matched-ratio PPL advantage persists across all five settings.

\end{document}